%% file: main.tex
\pdfoutput=1
\RequirePackage{fix-cm} 
\PassOptionsToPackage{dvipsnames}{xcolor}
\PassOptionsToPackage{capitalise}{cleveref}
\documentclass[11pt,a4paper]{academicreport}

\setcitestyle{numbers,square,comma,sort&compress}
\input{preamble}
\usepackage{fontawesome5}
\usepackage{inconsolata} 
\usepackage{fancyhdr}
\usepackage{hyperref}
\hypersetup{
  pdftitle    = {Don't Sleep, Keep Watch: Proactive Perception for Computer-Use Agents},
  pdfauthor   = {Jiaqi Su, Jie-Jing Shao, Yichen Liu, Long-Kai Huang, Haiyan Yin},
  pdfsubject  = {Proactive perception for computer-use agents},
  pdfkeywords = {computer-use agents, GUI agents, proactive perception, System-1 models, OSWorld}
}
\title{
  \fontsize{20pt}{24.5pt}\selectfont
  \textbf{Don't Sleep, Keep Watch: \textcolor{accent}{Proactive Perception}\\for Computer-Use Agents}
}
\author{
  \small
  \textbf{Jiaqi Su}\textsuperscript{1},\quad
  \textbf{Jie-Jing Shao}\textsuperscript{2},\quad
  \textbf{Yichen Liu}\textsuperscript{3},\quad
  \textbf{Long-Kai Huang}\textsuperscript{4},\quad
  \textbf{Haiyan Yin}\textsuperscript{2}\\[0.45em]
  \textsuperscript{1}National University of Singapore \quad
  \textsuperscript{2}A*STAR Centre for Frontier AI Research (A*STAR CFAR)\\
  \textsuperscript{3}SenseTime \quad
  \textsuperscript{4}Hong Kong Baptist University
}
\date{}

\fancypagestyle{firstpage}{
  \fancyhf{}
  \fancyhead[L]{\includegraphics[height=24pt]{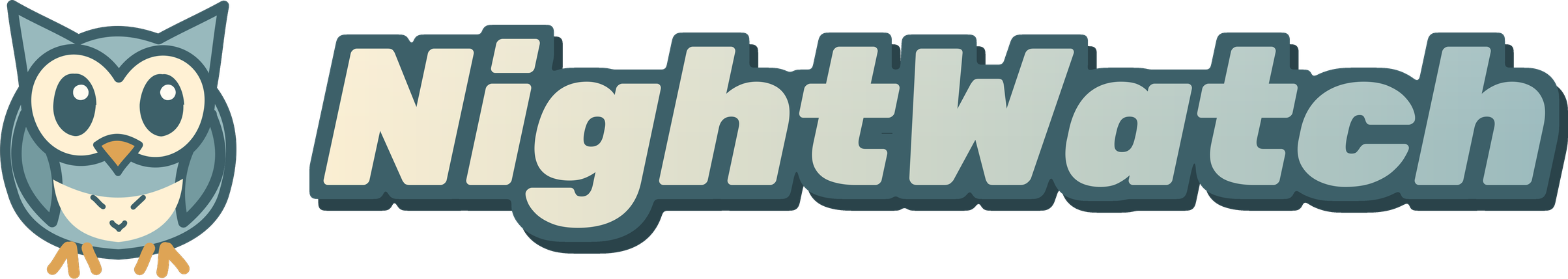}}
  \fancyhead[R]{
    \includegraphics[height=24pt]{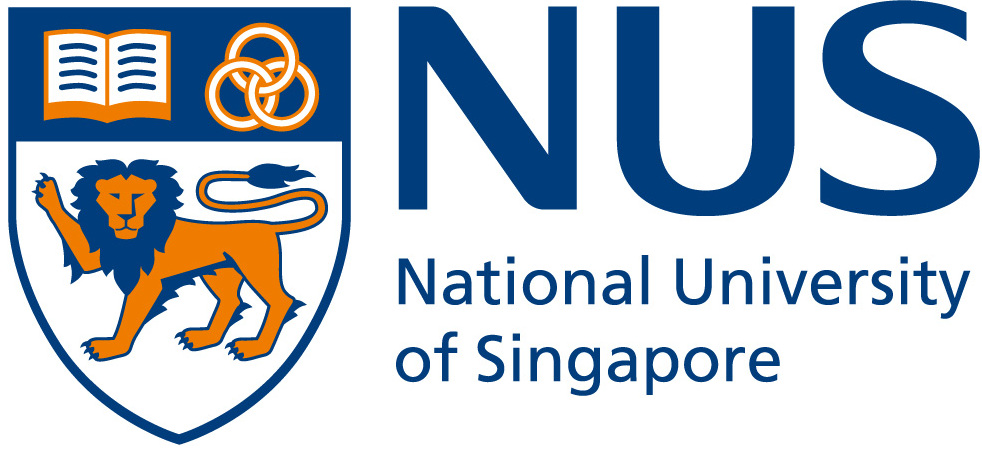}\hspace{1.1em}
    \raisebox{2pt}{\includegraphics[height=26pt]{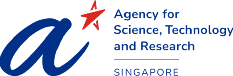}}
  }
  \fancyfoot[C]{\thepage}
  
}

\begin{document}

\maketitle
\thispagestyle{firstpage}
\vspace{-0.6em}
\input{sec/0_abstract}

\input{sec/1_intro}
\input{sec/2_related}
\input{sec/3_method}
\input{sec/4_experiments}
\input{sec/5_conclusion}


\begingroup
\renewcommand{\bibfont}{\footnotesize}
\bibliographystyle{assets/plainnat}
\bibliography{main}
\endgroup

\clearpage
\input{sec/X_suppl}

\end{document}

%% file: preamble.tex
\newcommand{\method}{\mbox{\textsc{NightWatch}}\xspace}
\newcommand{\watcher}{\mbox{\textsc{NightWatch-0.8B}}\xspace}

\usepackage{pifont}

\usepackage{colortbl}
\usepackage{multirow}
\usepackage{tabularx}

\newcommand{\ind}{\mathbf{1}}
\definecolor{bestcell}{HTML}{D3E5F8}
\definecolor{highlightblue}{HTML}{DCEBEA} 
\definecolor{secondcell}{HTML}{E9E9E9}
\definecolor{groupcell}{HTML}{EFEFEF}
\definecolor{ourscell}{HTML}{E8EAF8}
\definecolor{gaingreen}{HTML}{1E8E3E}
\newcommand{\gain}[1]{\textsubscript{\textcolor{gaingreen}{\textbf{($-$#1\%)}}}}

\makeatletter
\DeclareRobustCommand\onedot{\futurelet\@let@token\@onedot}
\def\@onedot{\ifx\@let@token.\else.\null\fi\xspace}
\def\eg{\emph{e.g}\onedot} 
\def\ie{\emph{i.e}\onedot} 
 
 \def\vs{\emph{vs}\onedot}

\makeatother

\usepackage{wrapfig}

\newcommand{\tact}[1]{t_{#1}^{\mathrm{act}}}
\newcommand{\trdy}[1]{t_{#1}^{\mathrm{rdy}}}
\definecolor{intcolor}{HTML}{AD493F}
\definecolor{voidcolor}{HTML}{A77620}
\definecolor{wakecolor}{HTML}{477A49}
\newcommand{\Int}{\textcolor{intcolor}{\textbf{Interrupt}}}
\newcommand{\Void}{\textcolor{voidcolor}{\textbf{Void}}}
\newcommand{\Wake}{\textcolor{wakecolor}{\textbf{Wake}}}

%% file: sec/0_abstract.tex
\begin{abstract}
Computer-use agents (CUAs) built on vision--language models are increasingly entrusted with long-horizon tasks in live desktop, web, and mobile environments, where the screen keeps changing on its own.
However, current CUAs lack real-time situational awareness: they perceive the environment through discrete snapshots, and the substantial inference latency of VLMs, compounded by highly dynamic interfaces, opens a perception--action gap in which the observed state frequently becomes stale before the corresponding action is executed.
To bridge this gap, we present \textbf{\method}, a plug-and-play framework that endows any CUA with continuous, temporally grounded perception without changing the agent's model, prompt, or action space.
\method couples the VLM agent with \watcher, a lightweight System-1 model that performs streaming monitoring of the interface state and keeps the agent synchronized with the evolving environment at low computational cost and latency, thereby grounding each decision in the current state rather than an outdated snapshot and reducing both stale actions and idle waiting while preserving task success.
Experiments on desktop and dynamic web environments (OSWorld-Verified and DynaCU-Bench) show that \method, driven by a watcher of only 0.8B parameters, reduces non-reasoning wall-clock time by 33\% and wasted time by 53\% on OSWorld while keeping task success on par, raises success on DynaCU-Bench from 39.5\% to 44.2\% with the large VLM agent, and is more efficient than passive-watching baselines such as AOI and PUSV under the same screenshot-only input, as well as a System-1 decision model more than $30\times$ its size.
Overall, \method demonstrates that the timing of perception is as critical as its content, marking a step toward agents that maintain continuous awareness of their environment rather than perceiving it turn by turn.
\par\medskip
{\noindent\small
\renewcommand{\arraystretch}{1.15}
\begin{tabular}{@{}c@{\hspace{0.6em}}l@{\hspace{0.6em}}l@{}}
\faGlobe & \textbf{\textcolor{accent}{Project Page:}} &
  \href{https://sujq0722.github.io/NightWatch/}{\nolinkurl{https://sujq0722.github.io/NightWatch}} \\
\faGithub & \textbf{\textcolor{accent}{Code:}} &
  \href{https://github.com/SuJQ0722/NightWatch}{\nolinkurl{https://github.com/SuJQ0722/NightWatch}} \\
\raisebox{-0.15\height}{\includegraphics[height=1.05em]{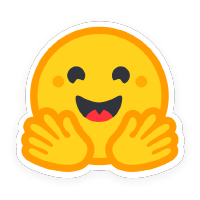}} &
  \textbf{\textcolor{accent}{Model:}} &
  \href{https://huggingface.co/nullptr04/NightWatch-0.8B}{\nolinkurl{https://huggingface.co/nullptr04/NightWatch-0.8B}} \\
\raisebox{-0.15\height}{\includegraphics[height=1.05em]{assets/huggingface_logo.png}} &
  \textbf{\textcolor{accent}{Data:}} &
  \href{https://huggingface.co/datasets/nullptr04/NightWatch-Data}{\nolinkurl{https://huggingface.co/datasets/nullptr04/NightWatch-Data}}
\end{tabular}
\par}
\end{abstract}

%% file: sec/1_intro.tex
\section{Introduction}
\label{sec:intro}

Computer-use agents (CUAs) built on vision--language models (VLMs) now operate real desktops, browsers, and phones~\cite{xie2024osworld,rawles2025androidworld,qin2025uitars,yuan2026osworld2}.
Unlike a single image, these screens move on their own: pages load, dialogs pop up, notifications flash and vanish, and animations settle on schedules that the agent does not control.
Yet almost every CUA still perceives its world through the same loop: take a screenshot, reason, act, sleep for a fixed delay, and repeat.

\begin{wrapfigure}{r}{0.56\linewidth}
    \centering\vspace{-1.2em}
    \includegraphics[width=\linewidth]{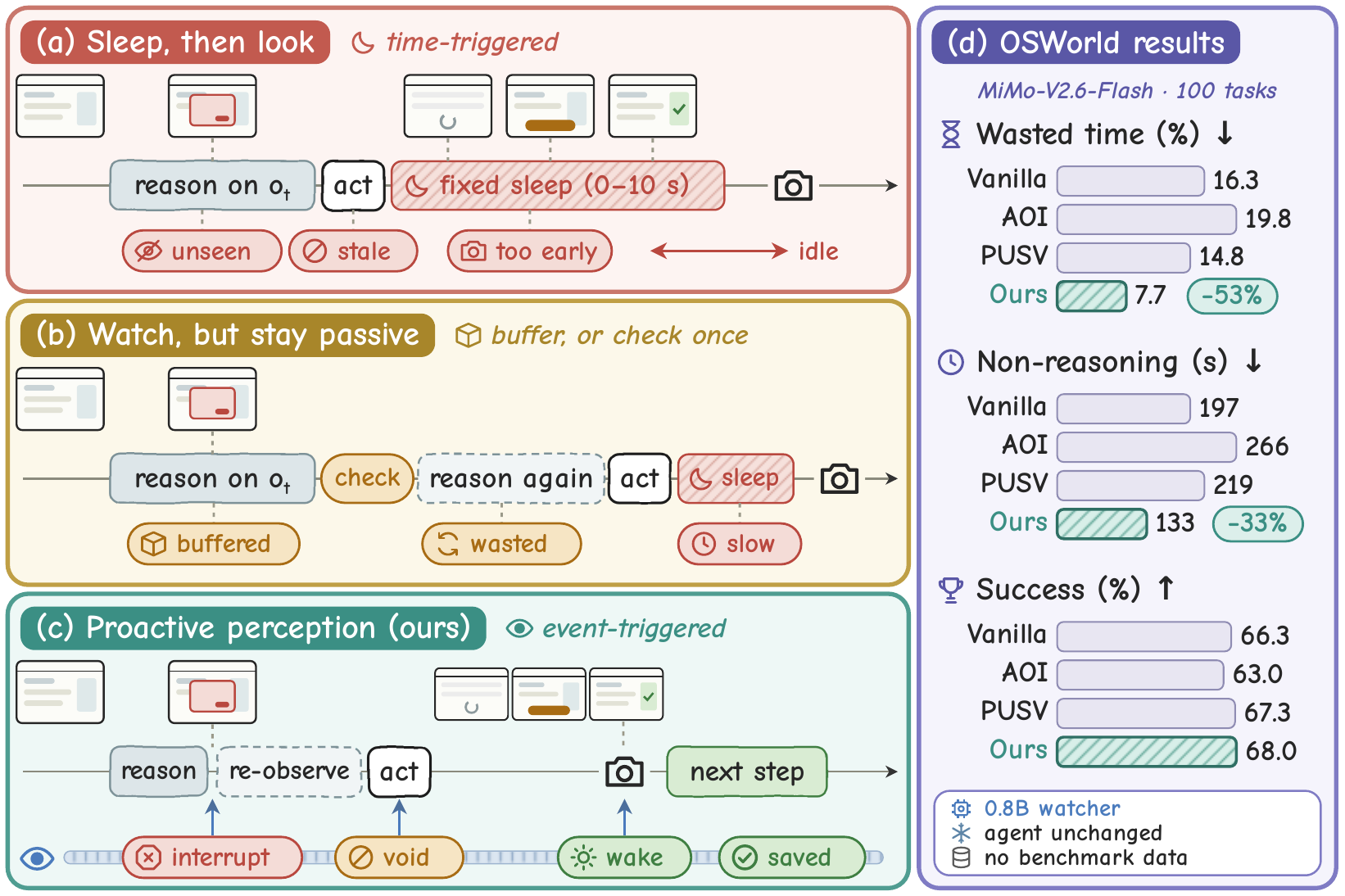}
    \caption{Observation and execution control in computer-use agents.
    \textbf{(a) Fixed scheduling.} Reasoning continues after a dialog appears, and fixed waits continue after the page is ready.
    \textbf{(b) Passive watching.} Buffered frames inform later observations, while pre-dispatch checks can reject actions after reasoning completes.
    \textbf{(c) Proactive perception.} \method combines a \textbf{0.8B System-1 watcher} with threshold rules to interrupt reasoning, void pending actions, and end post-action waits based on estimated interface state.
    \textbf{(d) OSWorld-Verified results.} \method reduces non-reasoning time by 33\% and the wasted-time share from 16.3\% to 7.7\% at comparable success with the same agent.}
    \label{fig:teaser}
\end{wrapfigure}

This loop is \emph{\textbf{reactive}}, or time-triggered: the agent looks only when its own clock says so, which has two consequences.
First, the agent is blind between screenshots.
The gap from observation to execution averages $6.51$\,s on OSWorld~\cite{xu2026toctou} and reaches $4.18$--$15.43$\,s on Android~\cite{qian2026mindthegap}.
A pop-up arriving within this gap can lower task success by $47\%$~\cite{zhang2025popups}, and agents that act on an outdated screen rarely recover from it~\cite{zhang2026believeeyes}.
Second, how long to wait is set by hand, with no principled signal that the screen is ready.
Across the official agent scripts of OSWorld~\cite{xie2024osworld}, the post-action sleep ranges from $0$ to $10$\,s, a twentyfold spread on a single benchmark.
A short sleep captures a half-rendered transition, whereas a long one idles while transient feedback disappears (\cref{fig:teaser}a).
Static benchmarks largely hide these failures behind generous waits~\cite{zhang2026cuadebug}, paying for them in wall-clock time that existing efficiency analyses attribute to model calls and step counts but not to sleep~\cite{abhyankar2026osworldhuman}.

Recent work has begun to watch the screen between steps, but the watching remains passive (\cref{fig:teaser}b).
AOI~\cite{li2026aoi} samples frames between steps yet only attaches them to the next scheduled observation, and PUSV~\cite{xu2026toctou} checks the screen once, right before dispatch. Event-triggered waits and monitors read DOM or trajectory text rather than pixels~\cite{maldaner2026sentinelbench,wei2026steplevel}, while pre-compiled policy trees reuse the same large VLM as their observer~\cite{dong2026aapt}.
None of them lets the screen decide when the agent looks, and none can stop reasoning that the screen has already made obsolete.

We argue that CUAs need \emph{\textbf{proactive perception}}: the perception layer, rather than the agent's fixed schedule, decides when the agent observes, and whether its ongoing reasoning and pending action are still valid.
The agent still pursues the task it was given, but no longer waits for its own clock to notice that the world has changed.
Proactive perception is close in spirit to streaming video models that decide when to respond without being asked~\cite{chen2024videollmonline,wang2025mmduet,qian2025dispider}, and to event-triggered control, which samples a system when its state demands rather than at fixed intervals~\cite{astrom2002riemann,tabuada2007event}.
Realizing it is hard: judgments must arrive in time, must separate changes that affect the pending action from unrelated animations, and err both ways, as a needless interrupt discards finished reasoning and an early wake shows an unready screen.

To this end, we present \textbf{\method}, a plug-and-play framework that equips any CUA with proactive perception.
A lightweight watcher continuously looks at the screen and answers a fixed set of event questions, and deterministic rules turn its answers into three interventions (\cref{fig:teaser}c): \emph{\textbf{interrupting}} reasoning that a screen change has made obsolete, \emph{\textbf{voiding}} a pending action whose target has gone stale, and \emph{\textbf{waking}} the agent the moment the screen settles, in place of the fixed sleep.
System-1 decision models~\cite{kahneman2011thinking,almeida2026jev,yu2026visualjev} are cheap enough to watch every frame, because they answer many multiple-choice questions from one encoding instead of generating text.
Recent work has begun to hand agent decisions to such models, which choose the next tool or action in place of the large model~\cite{zhu2026jevagentharness}.
Our small model instead never chooses what the agent does, but watches the screen and tells the agent when to look and whether its current reasoning or planned action still holds.
To our knowledge, \method is the first framework to bring proactive perception to CUAs, letting the screen decide when the agent looks and whether its reasoning and pending action still hold.

The design rests on three further observations.
The questions that matter, such as whether a dialog has appeared or whether the screen is still loading, are generic across applications, so the questions and their thresholds can be fixed once and reused across applications.
In addition, small models read states reliably but decide poorly, so the watcher only reads the screen while code makes the decision.
Finally, a quiet screen is not necessarily a settled one, since a page may pause while it loads, so the watcher learns readiness from hindsight labels, read from what the screen does next in recorded trajectories, rather than from stillness alone.
Our watcher, \watcher, is trained on desktop interaction trajectories from independently generated tasks and on procedurally generated synthetic interfaces, without any benchmark task, file, or trajectory.

With screenshot-only input on 100 OSWorld-Verified tasks, \method reduces non-reasoning wall-clock time by $33\%$ and wasted time by $53\%$ with an unchanged large VLM agent, while task success stays on par with the unmodified agent ($68.0\%$ vs.\ $66.3\%$). On the 49 dynamic web tasks of DynaCU-Bench, the same frozen watcher raises success from $39.5\%$ to $44.2\%$ with the large VLM agent and keeps it with the small one. Moreover, \method is more efficient than passive-watching baselines such as AOI~\cite{li2026aoi} and PUSV~\cite{xu2026toctou}, and in a separate comparison of watchers, \watcher is faster and more successful than a System-1 decision model more than $30\times$ its size.

Our contributions are as follows.
\begin{itemize}
    \item We identify reactive perception as an overlooked bottleneck of CUAs, formulate proactive perception, and quantify the cost of the time-triggered loop through a wall-clock ledger of idle waiting, premature observations, and stale actions (\cref{sec:formulation}).
    \item We propose \method, a plug-and-play framework in which a cheap watcher interrupts, voids, and wakes an unmodified agent, and train \watcher without benchmark data, releasing the code, data, and weights.
    \item Experiments on desktop and dynamic web environments show that \method saves $33\%$ of non-reasoning time on OSWorld-Verified with success on par, raises success by $4.7$ points on DynaCU-Bench with the large VLM agent, and that a small trained watcher is both faster and more successful than a decision model more than $30\times$ its size and a closed System-1 API.
\end{itemize}

%% file: sec/2_related.tex
\section{Related Work}
\label{sec:related}

\paragraph{Perception in computer-use agents.}
Benchmarks such as OSWorld~\cite{xie2024osworld,yuan2026osworld2} and AndroidWorld~\cite{rawles2025androidworld} drive VLM agents~\cite{qin2025uitars} through a discrete observe--reason--act loop, with a fixed pause after every action.
A growing body of work shows that this loop breaks once the screen changes on its own.
Pop-ups and environmental distractions mislead agents~\cite{zhang2025popups,ma2025caution}, and injected anomalies, notifications, and fake loading pages cause large drops in success~\cite{chen2026dgara,wu2026mobilebenchol,liu2026androidreality,wu2026realguinoise}.
Single screenshots miss intermediate states of dynamic interfaces~\cite{liu2026dynamicgui,yao2026livingscreen}, agents trust outdated structure over current pixels~\cite{zhang2026believeeyes}, and the delay between observing and acting opens a window for stale or even adversarial actions~\cite{xu2026toctou,qian2026mindthegap}, which time-sensitive and real-time agent benchmarks also penalize~\cite{froger2026gaia2,wen2026agilethinker}.
Several methods address parts of the problem.
AOI~\cite{li2026aoi} attaches keyframes sampled between steps to the next observation, whereas AAPT~\cite{dong2026aapt} pre-compiles a policy tree while the screen is idle and executes branches when changed frames match them. PUSV~\cite{xu2026toctou} validates the screen before dispatch, SentinelBench~\cite{maldaner2026sentinelbench} replaces sleeps with waits on DOM changes, and step-level monitors escalate to a stronger model on stuck or milestone events read from text~\cite{wei2026steplevel}.
Action-induced visual differences have also been used for reflection~\cite{ma2026efr}, and GUI testing has long replaced fixed waits with rendering-completion detectors~\cite{feng2023adat}.
Beyond GUIs, streaming video models decide frame by frame \emph{\textbf{when}} to respond~\cite{chen2024videollmonline,wang2025mmduet,qian2025dispider,ding2025streammind,wang2025streambridge,zhang2025proassist,zheng2026emgarde}, as rewarded by benchmarks of timely answers~\cite{li2025ovobench,lin2024streamingbench}, but learn this gate end to end inside the model.
\method instead leaves the agent untouched: a cheap visual watcher in front of it decides, from the pixels and at every frame, when the agent looks, and can interrupt reasoning or void actions instead of only buffering or checking once.

\paragraph{Fast watchers for slow reasoners.}
Pairing a fast monitor with a slow planner is common in robotics, where cheap checks of constraints or progress trigger replanning by a VLM~\cite{zhou2025codeasmonitor,guo2024doremi,agia2024sentinel,shi2025hirobot}, and real-time execution overlaps inference with action~\cite{black2025rtc,ramstedt2019rtrl,xiao2020thinking}.
Asynchronous LLM agents let events interrupt or redirect ongoing generation~\cite{gim2024asynclm,yakushev2026asyncllm,sun2026secondthought}.
System-1 decision models answer many typed questions from a single shared encoding at a fraction of the cost of a VLM call~\cite{yu2026visualjev}, and have been used as fast executors for mobile agents~\cite{zhang2026jevmobile}.
These ideas echo event-triggered control, which samples a system when its state demands rather than periodically~\cite{astrom2002riemann,tabuada2007event,heemels2012introduction}, and active perception, which treats when and how to sense as part of the task~\cite{bajcsy1988active}.
\method brings them to computer use, with a visual System-1 as the watcher and fixed thresholds that turn its answers into decisions.

%% file: sec/3_method.tex
\section{Method}
\label{sec:method}

\begin{figure}[!htb]
    \centering
    \includegraphics[width=\linewidth]{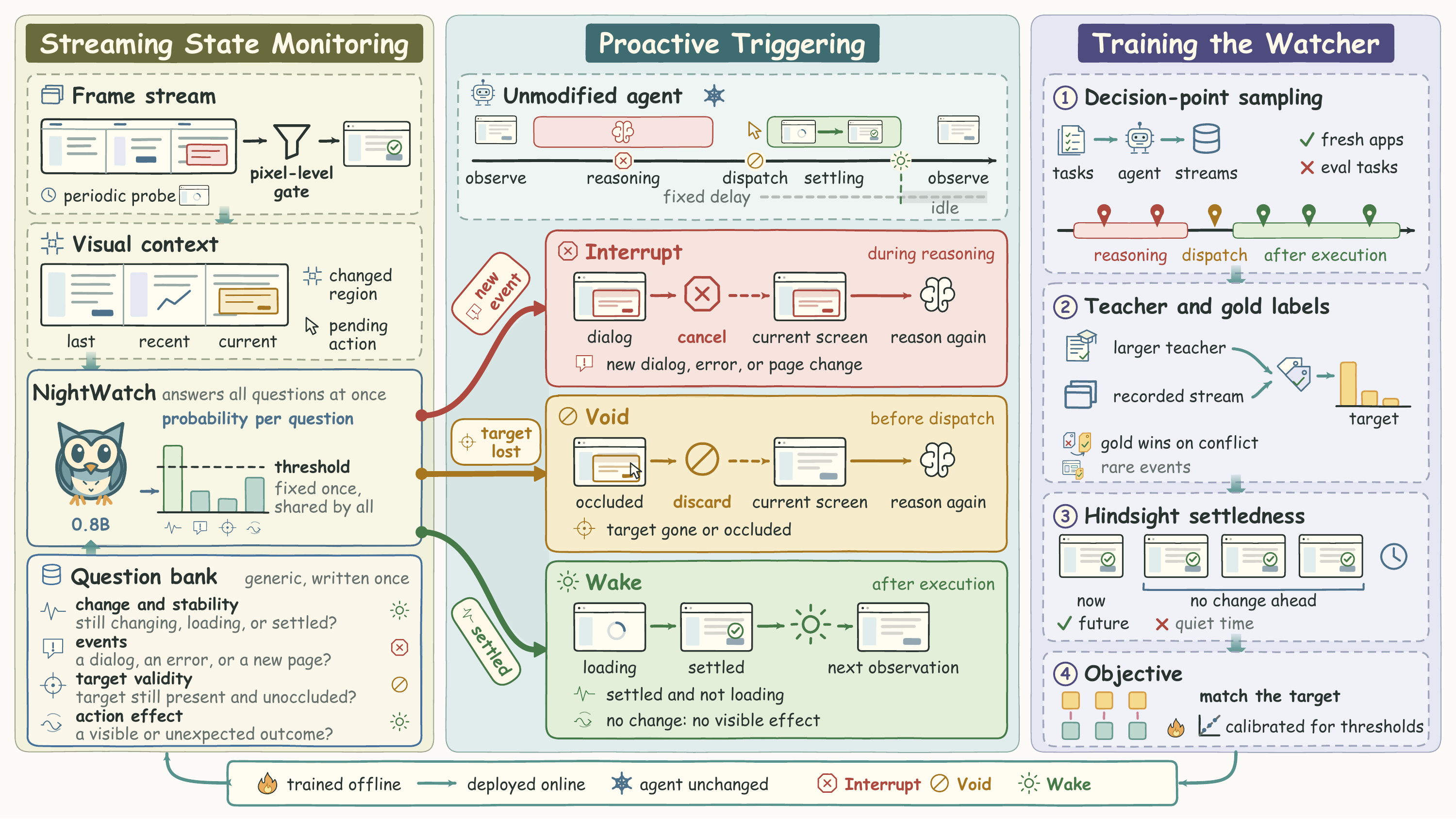}
    \caption{\textbf{Overview of \method.}
    \textbf{Left:}~the screen is watched as a stream, and \watcher reads a visual context built from recent frames and the pending action and answers every question of a fixed, generic bank with a probability.
    \textbf{Middle:}~in proactive triggering, fixed thresholds turn these probabilities into three interventions on an unmodified agent: \emph{\textbf{interrupt}} its reasoning when a new event appears, \emph{\textbf{void}} a pending action whose target is lost, and \emph{\textbf{wake}} it as soon as the screen settles.
    \textbf{Right:}~\watcher is trained offline on decision points from the agent's trajectories, with teacher, gold, and hindsight labels and a calibration-preserving objective.}
    \label{fig:method}
\end{figure}

\method operates at the harness level, placing a lightweight watcher between an unmodified VLM agent and its environment (\cref{fig:method}).
The agent keeps its own prompts, action space, and reasoning, while the watcher only decides \textbf{when} the agent observes and \textbf{whether} its ongoing reasoning and pending action remain valid.

\subsection{Problem Formulation}
\label{sec:formulation}

Let $s(t)$ be the screen at time $t$, which changes both in response to the agent's actions and on its own, and let $F_{\le t}$ be the stream of frames up to $t$.
An agent with policy $\pi$ proceeds in steps.
At step $k$ it observes $o_k=s(t_k)$ at time $t_k$, reasons, and dispatches an action $a_k\sim\pi(\cdot\mid o_k)$ at time $\tact{k}\ge t_k$ (the interaction history is omitted for brevity).
The screen may change both while the agent reasons, in $[t_k,\tact{k})$, and while the action executes and the agent waits, in $[\tact{k},t_{k+1})$.
The conventional, time-triggered loop ignores both and observes again after a fixed delay $t_{\text{fix}}$, \ie, $t_{k+1}=\tact{k}+t_{\text{fix}}$, which causes two timing failures.

\paragraph{Stale actions.}
The action is grounded in $o_k$ but applied to $s(\tact{k})$.
With a validity indicator $v(a,s)\in\{0,1\}$ of an action $a$ on a screen $s$, the action is \emph{\textbf{stale}} when $v(a_k,o_k)>v(a_k,s(\tact{k}))$, \ie, it was valid when generated, and the screen, not the agent, invalidated it.

\paragraph{Premature and late observations.}
The interface also needs time to become ready for the next decision.
Let $\trdy{k}\ge\tact{k}$ be the time at which the transition caused by $a_k$ is complete.
Readiness is semantic: it neither requires every pixel to stop changing, since a clock or a cursor may keep moving, nor follows from momentary stillness, since a page may pause while it loads.
An observation with $t_{k+1}<\trdy{k}$ is \emph{\textbf{premature}} and captures a transient state, whereas $t_{k+1}>\trdy{k}$ leaves the agent idle for $t_{k+1}-\trdy{k}$.

\paragraph{Proactive perception.}
We therefore make the observation times a decision of the harness.
A perception policy $\phi$, plugged in between the agent and its environment, reads $F_{\le t}$ and decides when the agent perceives the screen, and hence which screens it reasons on and acts against.
At step $k$, $\phi$ decides whether to interrupt the reasoning at some $t\in[t_k,\tact{k})$ and restart the step from $o_k=s(t)$, whether to void $a_k$ at $\tact{k}$, and when to take the next observation $t_{k+1}$.
The time-triggered loop is the special case $\phi_{\text{fix}}$ that never interrupts or voids and waits $t_{\text{fix}}$.
Under any $\phi$, the time wasted at step $k$ is the idle time plus a cost $c$ for each step spent on a premature observation or a stale action,
\begin{equation}
    W_k=\bigl(t_{k+1}-\trdy{k}\bigr)^{+}+c\,\Bigl(\ind\bigl[t_{k+1}<\trdy{k}\bigr]+\ind\bigl[v(a_k,o_k)>v\bigl(a_k,s(\tact{k})\bigr)\bigr]\Bigr),
    \label{eq:waste}
\end{equation}
and, with the agent fixed, we seek
\begin{equation}
    \min_{\phi}\ \mathbb{E}\Bigl[\sum_k W_k\Bigr]
    \quad \text{s.t.} \quad
    \mathrm{Succ}(\pi,\phi)\ \ge\ \mathrm{Succ}(\pi,\phi_{\text{fix}}),
    \label{eq:objective}
\end{equation}
which states the trade-off we target rather than an objective we optimize directly.
Neither $\trdy{k}$ nor $v$ can be read off when a decision is due, so $\phi$ must estimate them from $F_{\le t}$ as it unfolds: a learned watcher estimates them (\cref{sec:monitoring}), and fixed thresholds trigger the three decisions (\cref{sec:interventions}).

\subsection{Streaming State Monitoring}
\label{sec:monitoring}

The watcher is a System-1 decision model~\cite{kahneman2011thinking,yu2026visualjev} that answers typed questions about the interface without generating a reasoning trace.
For a question $q$ in a fixed bank $\mathcal{Q}$ with answer options $\mathcal{Y}_q$, it reads a context $x_t$ built at time $t$ and returns a distribution over the answers,
\begin{equation}
    p_\theta(y\mid x_t,q),\ \ y\in\mathcal{Y}_q,
    \qquad
    x_t=\Bigl(\underbrace{o_k,\ s(t-0.5\,\mathrm{s}),\ s(t),\ s(t)\vert_{B_t}}_{\text{visual context }u_t\text{ from }F_{\le t}},\ \underbrace{\bar a_t}_{\text{action}}\Bigr),
    \label{eq:watcher}
\end{equation}
where the visual context holds the agent's last observation, a recent frame, the current frame, and a crop of the region $B_t$ that changed, and $\bar a_t$ is the text of $a_k$ before dispatch or after execution and is empty while $a_k$ is still being generated (before dispatch, the crops cover the target of $a_k$; \cref{tab:supp_roles}).
We write $p_q(t)$ for the probability of the answer that fires $q$ (\eg, ``yes'' for a dialog, ``no'' for a present target) in the latest request completed by $t$.
The target-validity questions thus estimate whether $v(a_k,s(t))=0$, the settledness and loading questions whether $t\ge\trdy{k}$, and the event questions flag changes that make any plan obsolete.

\paragraph{Question bank.}
$\mathcal{Q}$ contains four groups of generic questions:
\ding{172}~\emph{\textbf{change and stability}}, \eg, whether the screen is still changing or loading, or has settled;
\ding{173}~\emph{\textbf{events}}, \eg, whether a dialog, an error, or a different page has appeared;
\ding{174}~\emph{\textbf{target validity}}, \eg, whether the element the agent is about to act on is still present and unoccluded;
\ding{175}~\emph{\textbf{action effect}}, \eg, whether the last action produced a visible or an unexpected outcome.
$\mathcal{Q}$ is \emph{\textbf{generic}} by design: no question names an application or a task, so the bank and the thresholds of \cref{sec:interventions} are fixed once rather than generated per step (all 16 questions are in \cref{sec:supp_bank}, and the exact request format in \cref{sec:supp_prompts}).

\paragraph{Streaming inference.}
A pixel-level gate sets $\chi(t)=1$ when frame $s(t)$ differs significantly from its predecessor and requests an evaluation only then, while a periodic probe keeps a static loading screen under examination.
Each request asks the subset $\mathcal{Q}_t\subseteq\mathcal{Q}$ of the current stage and encodes the visual context once: with a visual encoder $E_\theta$ and an answer readout $R_\theta$,
\begin{equation}
    p_\theta(\cdot\mid x_t,q)=R_\theta\bigl(E_\theta(u_t),\,\bar a_t,\,q\bigr)\quad\text{for every } q\in\mathcal{Q}_t,
    \label{eq:stream}
\end{equation}
so the visual cost is paid once per request rather than once per question.
While the agent waits, $\mathcal{Q}_t=\{\textit{settled},\textit{loading}\}$ is served apart from the full checks to keep waiting short.

\subsection{Proactive Triggering}
\label{sec:interventions}

Proactive triggering lets the screen, not the agent's clock, fire each decision of $\phi$.
Fixed rules compare the watcher's probabilities with thresholds, so the model only reads the state while the decision logic stays transparent.
For a set of questions $S\subseteq\mathcal{Q}$, let $g(S,t)=\max_{q\in S}p_q(t)$ be the strongest response among them.
Each intervention $d\in\{I,V,W\}$ watches a time window $\mathcal{W}_d$ of step $k$ and fires at the first moment its trigger $T_d(t)\in\{0,1\}$ holds,
\begin{equation}
    t_d^{\star}=\min\bigl\{\,t\in\mathcal{W}_d : T_d(t)=1\,\bigr\},
    \label{eq:trigger}
\end{equation}
and the three interventions differ only in their window, trigger, and effect.

\paragraph{\textcolor{intcolor}{Interrupt.}}
While the agent reasons on $o_k$, the event questions $S_{\text{evt}}$ (a new dialog, an error, or a page change) are asked about every significant change of the screen:
\begin{equation}
    \mathcal{W}_I=[t_k,\ \tact{k}),\qquad
    T_I(t)=\ind\bigl[g(S_{\text{evt}},t)\ge\tau_{I}\bigr].
    \label{eq:interrupt}
\end{equation}
At $t_I^{\star}$, the generation is cancelled and step $k$ restarts from $o_k=s(t_I^{\star})$.

\paragraph{\textcolor{voidcolor}{Void.}}
Before dispatch, the target-validity and dialog questions $S_{\text{val}}$ are asked about the target of $a_k$:
\begin{equation}
    \mathcal{W}_V=\{\tact{k}\},\qquad
    T_V(t)=\ind\bigl[g(S_{\text{val}},t)\ge\tau_{V}\bigr],
    \label{eq:void}
\end{equation}
so that $T_V$ estimates whether $v(a_k,s(\tact{k}))=0$.
If it fires, $a_k$ is discarded rather than executed on an outdated screen, and step $k$ restarts from $o_k=s(t_V^{\star})$.

\paragraph{\textcolor{wakecolor}{Wake.}}
After execution, the wake replaces the fixed delay $t_{\text{fix}}$ with an online estimate of the ready time $\trdy{k}$, made in two stages:
\begin{equation}
    \mathcal{W}_W=[\tact{k},\ \infty),\qquad
    T_W(t)=\ind\bigl[\eta(t)\ge t_{\min}\bigr]\cdot\ind\bigl[p_{\text{settled}}(t)\ge\tau_{W}\bigr]\cdot\ind\bigl[p_{\text{loading}}(t)<\tau_{L}\bigr],
    \label{eq:wake}
\end{equation}
and the next observation is taken at $t_{k+1}=t_W^{\star}$.
The gate first establishes that the screen has stopped changing, with $\eta(t)$ the time since the last $\chi=1$ after $\tact{k}$, or since $\tact{k}$ if there has been none, so that an action with no visible effect needs no timeout.
The watcher then judges whether this stillness is final, which pixels alone cannot tell (\cref{sec:formulation}).
If the watcher does not confirm, the agent is woken after a longer quiet time unless the watcher still reports the screen as busy, and a maximum waiting time bounds every step.
The wake thus reduces the idle term of \cref{eq:waste} to a detection delay, analyzed in \cref{sec:supp_wake}.

The time-triggered loop $\phi_{\text{fix}}$ is recovered with $T_{I}=T_{V}=0$ and $t^{\star}_{W}=\tact{k}+t_{\text{fix}}$.
The thresholds $\tau_{I},\tau_{V},\tau_{W},\tau_{L}$ and $t_{\min}$ are fixed once and shared across applications and benchmarks (\cref{sec:supp_bank}).

\subsection{Training the Watcher}
\label{sec:training}

A general-purpose decision model reads static states reliably but is weak at temporal judgments, such as whether a screen that has just stopped moving is final, so we train the watcher on the states that the agent visits, with labels from a teacher and from the frame stream in hindsight.

\paragraph{Training environments.}
The training tasks $\mathcal{T}_{\text{train}}$ use the operating system and applications of the evaluation, each starting from a freshly launched application, and tasks close to an evaluation task are discarded, so that no evaluation task, file, or trajectory is used.

\paragraph{Decision-point sampling.}
We run the unmodified agent $\pi$ on $\mathcal{T}_{\text{train}}$ while recording the frame stream.
From each trajectory we sample decision points at the three stages of \cref{sec:interventions}: frames during reasoning, the last frame before dispatch, and frames at several delays after execution.
Each decision point gives a context $x_i$ built exactly as at inference and a question $q_i$ asked at its stage, forming $\mathcal{D}=\{(x_i,q_i)\}$.

\paragraph{Teacher and gold labels.}
The target $y_i$ over $\mathcal{Y}_{q_i}$ combines the distribution $y_i^{\text{T}}$ of a larger teacher decision model on the same request with a gold label $y_i^{\text{G}}$ wherever the recorded stream determines the answer, \eg, no event when nothing changed during reasoning, or an unchanged target region before dispatch.
When both exist and agree on the answer,
\begin{equation}
    y_i=\alpha\, y_i^{\text{G}}+(1-\alpha)\, y_i^{\text{T}} .
    \label{eq:labels}
\end{equation}
Without a gold label, $y_i=y_i^{\text{T}}$; when the two disagree, a gold label that states a fact about the pixels, as in both examples above, replaces the teacher, and the teacher is kept otherwise.
Synthetic interfaces with exact ground truth complement the recordings for rare events such as dialogs, toasts, and error messages.

\paragraph{Hindsight settledness.}
For the settledness question, the gold label is defined by the future rather than the past:
\begin{equation}
    y_i^{\text{G}}=\ind\bigl[\,\chi(t')=0\ \ \forall\, t'\in(t_i,t_i+h]\,\bigr],
    \label{eq:hindsight}
\end{equation}
where $t_i$ is the time of the decision point and $t'$ ranges over the recorded frames, a hindsight proxy for the ready time $\trdy{k}$ of \cref{sec:formulation}.
A label based on elapsed quiet time would mark every frame that has just stopped moving as unsettled and could never wake earlier than a pixel-quiescence rule.
We sample these examples where the wake rule decides, at the first quiet frames after each change and at short delays after an action that has not yet changed the screen, and keep only windows that close before the next dispatch.

\paragraph{Objective.}
With $p_{\theta,i}=p_\theta(\cdot\mid x_i,q_i)$, we fit the watcher with a cross-entropy term and a Brier term over the answer options,
\begin{equation}
    \mathcal{L}(\theta)=\mathbb{E}_{i\sim\mathcal{D}}\bigl[\mathrm{CE}(y_i,p_{\theta,i})+\lambda\,\|y_i-p_{\theta,i}\|_2^2\bigr],
    \label{eq:loss}
\end{equation}
where the Brier term preserves the calibration that the thresholds in \cref{sec:interventions} rely on.
A final temperature is fitted on a held-out split of $\mathcal{T}_{\text{train}}$.

%% file: sec/4_experiments.tex
\section{Experiments}
\label{sec:experiments}

\subsection{Experimental Setup}
\label{sec:setup}

\paragraph{Evaluation benchmarks.}
We evaluate \method in two computer-use domains with different dynamics.
\ding{172}~\emph{\textbf{Desktop}}: OSWorld-Verified~\cite{xie2024osworld}, on a random sample of 100 tasks that covers all ten application domains.
\ding{173}~\emph{\textbf{Dynamic web}}: the 49 dynamic tasks of DynaCU-Bench~\cite{li2026aoi} that do not require audio, such as live dashboards, carousels, and games, in which the page changes while the agent acts.
Each task is run three times.
OSWorld is a largely static benchmark on which generous waits hide timing failures~\cite{zhang2026cuadebug}, so we expect savings in wall-clock time with success on par, whereas DynaCU-Bench tests whether the same watcher also helps success when the screen does not wait for the agent.

\paragraph{Baselines.}
All methods share the same agent, prompt, action space, and step budget, and differ only in how the agent perceives the screen.
We compare against two paradigms.
\begin{itemize}[leftmargin=*,itemsep=1pt,topsep=2pt]
    \item[$\blacksquare$] \emph{\textbf{Time-triggered}}: the official loop of each benchmark (\emph{\textbf{Vanilla}}), which observes after the fixed pauses of the benchmark's official setting, such as AOI's 2\,s interval on DynaCU-Bench.
    \item[$\blacksquare$] \emph{\textbf{Passive watching}}: AOI~\cite{li2026aoi}, which samples frames at 3\,Hz between steps and attaches pixel- and CLIP-gated keyframes with a visual narration to the next scheduled observation, and PUSV~\cite{xu2026toctou}, which verifies the screen around the target once before each click and lets the agent re-decide when the check fails.
\end{itemize}
\Cref{sec:ablation} compares other schemes and watchers.

\paragraph{Implementation.}
Our harness follows the official setting of each benchmark: the agent is OSWorld's screenshot-only prompt agent~\cite{xie2024osworld} on OSWorld and AOI's standard agent~\cite{li2026aoi} on DynaCU-Bench, each with two unchanged backbones: MiMo-V2.6-Flash~\cite{xiaomi2026mimov26}, a hosted large VLM, and Qwen3.8-27B (FP8)~\cite{qwen38}, a small open-weight VLM.
Tasks allow at most 100 executed steps on OSWorld and 50 on DynaCU-Bench, and all methods receive screenshots and the trajectory only, so AOI's DOM element list is disabled.
On DynaCU-Bench, all methods share AOI's harness, with its agent, prompt, action parsing, failure feedback, and judge, and each baseline is reproduced as described in its paper (\cref{sec:supp_protocol}).
\watcher is initialized from d3-lite~\cite{d3lite2026}, a 0.8B System-1 decision model, fine-tuned with LoRA on 360 generated OSWorld tasks, and used frozen, with the same thresholds, on both benchmarks.
It runs on one NVIDIA L40S GPU over a 10\,Hz frame stream, and a settledness check takes about 0.3\,s.
\cref{sec:supp_training,sec:supp_protocol} detail the training data, the harnesses, and the judge.

\paragraph{Metrics.}
We report task success averaged over runs (avg@3) and the share of tasks solved in any run (pass@3).
For efficiency, we report the \emph{\textbf{non-reasoning time}} per task, \ie, wall-clock time minus the time spent in agent LLM calls, which isolates the part of the loop that perception controls, and the \emph{\textbf{wasted-time share}}, the fraction of loop time spent idle, on premature observations, or on stale actions.
For timeliness, we report the share of steps with a premature observation or a stale action, as defined in \cref{sec:formulation} and measured from screen changes (\cref{sec:supp_metrics}).

\subsection{Main Results}
\label{sec:main}

\begin{table}[!htb]
\centering
\caption{\textbf{Main results} on OSWorld-Verified (100 tasks) and DynaCU-Bench (49 dynamic tasks) with two unmodified agents and screenshot-only input.
Succ.: success averaged over three runs (mean $\pm$ std); pass@3: share of tasks solved in any run; Non-reas.: wall-clock time per task outside agent LLM calls; Waste: share of loop time that is idle, premature, or stale; Prem./Stale: share of executed steps with a premature observation or a stale action. Green subscripts: relative reduction against Vanilla.
$\uparrow$/$\downarrow$: higher/lower is better. Within each agent block, the best result in each column is bold with a teal background, the second-best distinct value has a gray background, and ties receive the same formatting.
--: not measured.}
\label{tab:main}
\setlength{\tabcolsep}{4.2pt}
\renewcommand{\arraystretch}{1.2}
\small
\resizebox{\linewidth}{!}{%
\begin{tabular}{llcccccc ccc}
\toprule
& & \multicolumn{6}{c}{\textbf{OSWorld-Verified}} & \multicolumn{3}{c}{\textbf{DynaCU-Bench}} \\
\cmidrule(lr){3-8} \cmidrule(lr){9-11}
\textbf{Method} & \textbf{Paradigm} & Succ.$\uparrow$ & pass@3$\uparrow$ & Non-reas.\,(s)$\downarrow$ & Waste\,(\%)$\downarrow$ & Prem.\,(\%)$\downarrow$ & Stale\,(\%)$\downarrow$ & Succ.$\uparrow$ & pass@3$\uparrow$ & Non-reas.\,(s)$\downarrow$ \\
\midrule
\multicolumn{11}{c}{\textit{Large VLM agent: MiMo-V2.6-Flash}} \\
\midrule
Vanilla & Time-triggered & 66.3\,$\pm$\,4.5 & \cellcolor{highlightblue}\textbf{82.0} & \cellcolor{gray!15}197.4 & 16.3 & 9.6 & 17.4 & \cellcolor{gray!15}39.5\,$\pm$\,4.2 & \cellcolor{gray!15}53.1 & -- \\
AOI~\cite{li2026aoi} & Passive watching & 63.0\,$\pm$\,3.6 & 78.0 & 266.4 & 19.8 & \cellcolor{gray!15}3.5 & \cellcolor{gray!15}8.9 & 37.4\,$\pm$\,4.2 & \cellcolor{highlightblue}\textbf{55.1} & -- \\
PUSV~\cite{xu2026toctou} & Passive watching & \cellcolor{gray!15}67.3\,$\pm$\,0.6 & \cellcolor{gray!15}81.0 & 218.6 & \cellcolor{gray!15}14.8 & 7.1 & 12.2 & 33.3\,$\pm$\,4.2 & 49.0 & \cellcolor{gray!15}46 \\
\textbf{\method (Ours)} & Proactive watching & \cellcolor{highlightblue}\textbf{68.0\,$\pm$\,3.6} & \cellcolor{highlightblue}\textbf{82.0} & \cellcolor{highlightblue}\textbf{132.7}\gain{33} & \cellcolor{highlightblue}\textbf{7.7}\gain{53} & \cellcolor{highlightblue}\textbf{2.6}\gain{73} & \cellcolor{highlightblue}\textbf{8.3}\gain{52} & \cellcolor{highlightblue}\textbf{44.2\,$\pm$\,5.1} & \cellcolor{gray!15}53.1 & \cellcolor{highlightblue}\textbf{30} \\
\midrule
\multicolumn{11}{c}{\textit{Small VLM agent: Qwen3.8-27B (FP8)}} \\
\midrule
Vanilla & Time-triggered & \cellcolor{gray!15}56.3\,$\pm$\,2.1 & 69.0 & 190.2 & 16.2 & 11.4 & 17.2 & \cellcolor{highlightblue}\textbf{34.7\,$\pm$\,5.4} & \cellcolor{gray!15}46.9 & -- \\
AOI~\cite{li2026aoi} & Passive watching & \cellcolor{gray!15}56.3\,$\pm$\,1.2 & 74.0 & \cellcolor{gray!15}124.8 & \cellcolor{gray!15}7.6 & \cellcolor{gray!15}4.9 & \cellcolor{gray!15}12.7 & \cellcolor{gray!15}34.0\,$\pm$\,2.4 & 44.9 & -- \\
PUSV~\cite{xu2026toctou} & Passive watching & \cellcolor{highlightblue}\textbf{58.7\,$\pm$\,4.9} & \cellcolor{gray!15}75.0 & 165.8 & 13.1 & 9.2 & 13.2 & 32.0\,$\pm$\,6.2 & 44.9 & \cellcolor{gray!15}25 \\
\textbf{\method (Ours)} & Proactive watching & \cellcolor{highlightblue}\textbf{58.7\,$\pm$\,1.2} & \cellcolor{highlightblue}\textbf{77.0} & \cellcolor{highlightblue}\textbf{111.9}\gain{41} & \cellcolor{highlightblue}\textbf{5.4}\gain{67} & \cellcolor{highlightblue}\textbf{1.9}\gain{83} & \cellcolor{highlightblue}\textbf{11.8}\gain{31} & \cellcolor{highlightblue}\textbf{34.7\,$\pm$\,4.1} & \cellcolor{highlightblue}\textbf{49.0} & \cellcolor{highlightblue}\textbf{17} \\
\bottomrule
\end{tabular}}
\end{table}

\begin{figure}[!htb]
\centering
\includegraphics[width=\linewidth]{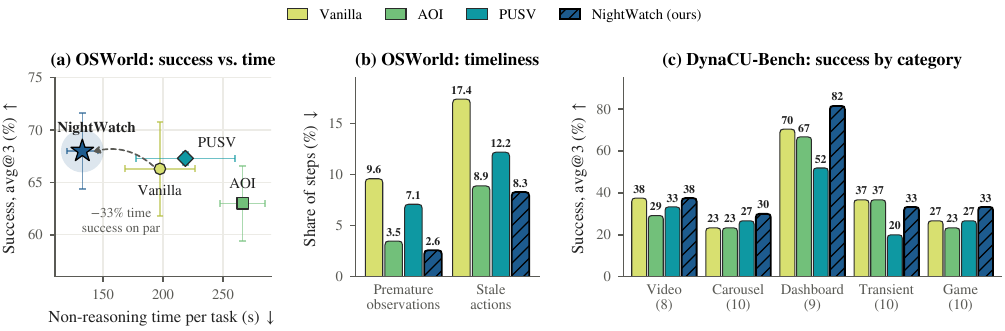}
\caption{\textbf{Results at a glance} for the large VLM agent.
(a)~Success against non-reasoning time per task on OSWorld, with error bars showing the standard deviation over three runs.
(b)~Share of steps with a premature observation or a stale action on OSWorld.
(c)~Success by task category on DynaCU-Bench (number of tasks in parentheses, with the two single-task categories omitted).}
\label{fig:results}
\end{figure}

\paragraph{Desktop.}
As \cref{tab:main} and \cref{fig:results}a show, \method spends the least time outside reasoning on OSWorld with both agents while keeping success on par with the strongest baseline.
For the large VLM agent, success is 68.0\% against 66.3\% for Vanilla and 67.3\% for PUSV, while non-reasoning time falls from 197.4\,s to 132.7\,s ($-33\%$) and the wasted-time share from 16.3\% to 7.7\% ($-53\%$).
For the small VLM agent, success matches PUSV (58.7\% against 56.3\% for Vanilla), pass@3 rises from 69.0\% to 77.0\%, and non-reasoning time falls from 190.2\,s to 111.9\,s ($-41\%$).
Both reductions in non-reasoning time are significant in a paired test over tasks ($p=0.026$ and $p=0.016$; \cref{sec:supp_stats}).
Passive watching does not act earlier: with the large VLM agent, AOI and PUSV spend 9.2\,s and 7.4\,s per step outside reasoning, against 6.9\,s for Vanilla and 4.5\,s for \method (\cref{sec:supp_results}).

\paragraph{Dynamic web.}
\method reaches 44.2\% with the large VLM agent on DynaCU-Bench, 4.7 points above Vanilla (39.5\%) and 6.8 above AOI (37.4\%), and spends 30\,s of non-reasoning time per task against 46\,s for PUSV; with the small VLM agent, it matches Vanilla (34.7\%) with 17\,s against 25\,s for PUSV.
PUSV falls below Vanilla with both agents (33.3\% and 32.0\%), mostly on live dashboards and transient notifications (\cref{fig:results}c), because a pixel-level check cannot tell a benign refresh from a change that invalidates the click.
AOI runs without its DOM list, so that all methods receive the same screenshot-only input (\cref{sec:supp_results}).

\paragraph{Timeliness.}
For the large VLM agent, \method cuts premature observations on OSWorld from 9.6\% to 2.6\% of executed steps ($-73\%$) and stale actions from 17.4\% to 8.3\% ($-52\%$; \cref{fig:results}b), and the small VLM agent follows the same trend (11.4\% to 1.9\% and 17.2\% to 11.8\%).
AOI comes second with both agents, partly because its agent waits longer after each action.

\paragraph{Per-category results.}
Across the categories of DynaCU-Bench, \method is best or tied for best on videos, carousels, dashboards, and games (\cref{fig:results}c).
Its largest margin appears on live dashboards (81.5\% \vs 70.4\% for Vanilla), where waking on a settled panel rather than on a fixed clock matters most.
\cref{sec:supp_failure} gives a failure analysis.

\subsection{Cost Efficiency}
\label{sec:cost}

\begin{figure}[!htb]
\centering
\includegraphics[width=0.85\linewidth]{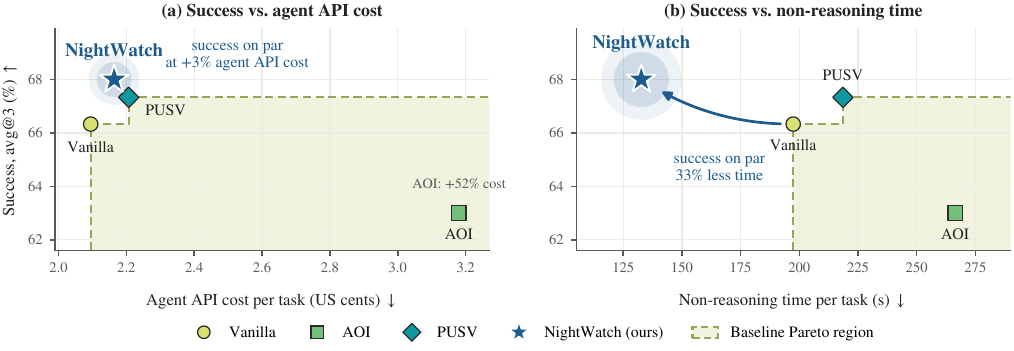}
\caption{\textbf{Success against agent API cost and non-reasoning time} on OSWorld-Verified with MiMo-V2.6-Flash (means over three runs; the standard deviations are in \cref{tab:main}). The shaded region is dominated by the baselines' Pareto front.}
\label{fig:pareto}
\end{figure}

End to end (\cref{tab:cost}), \method is the fastest method with both agents: a task takes 614.7\,s against 666.0\,s for Vanilla with the large VLM agent ($-8\%$) and 625.4\,s against 701.9\,s with the small one ($-11\%$).
The generations that it interrupts or voids cost only 2.5\,s and 4.5\,s per task, and it issues about as many agent calls as Vanilla, so the time saved outside reasoning is not spent on extra reasoning.
Its agent API cost stays within 3\% of Vanilla, whereas AOI's keyframes enlarge each call from about 4.5k to 7k input tokens and raise the cost by 52\% and 37\%.
The watcher adds a 0.8B model on one L40S GPU and no API calls.
With the large VLM agent, \method lies outside the region dominated by the baselines in both cost and time (\cref{fig:pareto}).

\begin{table}[!htb]
\centering
\caption{\textbf{End-to-end cost} per task on OSWorld-Verified (300 runs). Total: loop time (s); Disc.: discarded generation (s); API: agent API cost (US cents), counting interrupted calls.}
\label{tab:cost}
\small
\setlength{\tabcolsep}{4pt}
\begin{tabular}{lcccccc}
\toprule
& \multicolumn{3}{c}{\textbf{MiMo-V2.6-Flash}} & \multicolumn{3}{c}{\textbf{Qwen3.8-27B}} \\
\cmidrule(lr){2-4} \cmidrule(lr){5-7}
\textbf{Method} & Total$\downarrow$ & Disc.$\downarrow$ & API$\downarrow$ & Total$\downarrow$ & Disc.$\downarrow$ & API$\downarrow$ \\
\midrule
Vanilla & 666.0 & 0.0 & 2.10 & 701.9 & 0.0 & 1.48 \\
AOI & 789.1 & 0.0 & 3.18 & 686.1 & 0.0 & 2.02 \\
PUSV & 700.7 & 15.0 & 2.21 & 676.9 & 25.6 & 1.55 \\
\textbf{\method} & \textbf{614.7} & 2.5 & 2.16 & \textbf{625.4} & 4.5 & \textbf{1.45} \\
\bottomrule
\end{tabular}
\end{table}

\subsection{Ablation Study}
\label{sec:ablation}

\begin{table}[!htb]
\centering
\footnotesize
\setlength{\tabcolsep}{4pt}
\begin{minipage}[t]{0.47\linewidth}
\centering
\caption{\textbf{Control schemes} on DynaCU-Bench (MiMo-V2.6-Flash, three runs). Time and Non-reas.: s per task. Fixed wait re-implements Vanilla's 2\,s pause in our controller, in runs separate from \cref{tab:main}.}
\label{tab:watcher}
\begin{adjustbox}{max width=\linewidth}
\begin{tabular}{lccc}
\toprule
\textbf{Control scheme} & \textbf{Succ.}$\uparrow$ & \textbf{Time}$\downarrow$ & \textbf{Non-reas.}$\downarrow$ \\
\midrule
Fixed wait & 33.3 & 127.6 & 43.5 \\
Pixel rules & 30.6 & 132.0 & 33.1 \\
Jev API watcher & 36.7 & 128.9 & 33.6 \\
D3-27B (FP8) & 32.7 & 124.3 & 54.7 \\
\midrule
\textbf{\watcher (ours)} & \textbf{44.2} & \textbf{113.4} & \textbf{29.8} \\
\bottomrule
\end{tabular}
\end{adjustbox}
\end{minipage}\hfill
\begin{minipage}[t]{0.49\linewidth}
\centering
\caption{\textbf{Interventions} on DynaCU-Bench (three runs). Each variant removes one intervention; w/o wake waits for pixel quiescence. Mism.: targeted actions whose target changed before dispatch (\%); Fired: steps on which the removed intervention fires in \method (\%).}
\label{tab:interventions}
\begin{adjustbox}{max width=\linewidth}
\begin{tabular}{lcccc}
\toprule
\textbf{Variant} & \textbf{Succ.}$\uparrow$ & \textbf{Time}$\downarrow$ & \textbf{Mism.}$\downarrow$ & \textbf{Fired} \\
\midrule
\textbf{\method (full)} & \textbf{44.2} & \textbf{113.4} & \textbf{3.6} & -- \\
\midrule
\quad w/o wake & 38.1 & 122.7 & 3.8 & 49.0 \\
\quad w/o interrupt & 30.6 & 133.1 & 4.2 & 15.0 \\
\quad w/o void & 36.1 & 132.6 & 15.9 & 6.2 \\
\bottomrule
\end{tabular}
\end{adjustbox}
\end{minipage}
\end{table}

\paragraph{\ding{182}~Control schemes.}
\cref{tab:watcher} isolates the decision layer.
Fixed wait is the time-triggered loop, pixel rules make the same three decisions from pixel differences alone, and the Jev API watcher, a closed System-1 model~\cite{almeida2026jev} served through an API, answers the same questions as \watcher.
D3-27B is the official FP8 decision model~\cite{d32026} with the same question bank and thresholds, and its latency stays in the reported time (\cref{sec:supp_protocol}).
\watcher takes the least time per task, 113.4\,s against 124.3--132.0\,s for the other schemes, and the least time outside reasoning, 29.8\,s against 33.1--54.7\,s.
The comparison is end to end: D3-27B, more than $30\times$ larger, takes 7.6\,s per call, and the Jev API watcher 0.59\,s per check before dispatch against 0.25\,s.
With the same controller, d3-lite without fine-tuning spends 35.0\,s outside reasoning against 28.8\,s in the same runs.

\paragraph{\ding{183}~Interventions.}
\cref{tab:interventions} removes one intervention at a time on DynaCU-Bench, where all three fire often: \watcher wakes the agent early on about half of the steps, and interrupts and voids on 15 and 6 of every 100 steps.
Removing any intervention lengthens the task without raising success, interrupt the most (19.7\,s per task).
Removing void raises mismatched actions from 3.6\% to 15.9\%, and replacing wake with pixel quiescence adds 9.3\,s per task.

\paragraph{\ding{184}~Watcher against pixel quiescence on OSWorld.}
On OSWorld, where \method interrupts only 0.48 times and voids 0.03 actions per task, the wake carries most of the gain.
A pixel-only variant that uses the same harness but wakes the agent on pixel quiescence alone, and never interrupts or voids, spends 179.8\,s per task outside reasoning and 770.8\,s in total, against 113.1\,s ($-37\%$) and 611.5\,s ($-21\%$) for \method in a run with the same agent, and \method is faster on two thirds of the tasks (\cref{tab:supp_pixel}).
Judging whether stillness is final, rather than waiting for it to last, is thus what shortens the wait.

\subsection{Qualitative Analysis}
\label{sec:qualitative}

\begin{figure}[!htb]
\centering
\includegraphics[width=\linewidth]{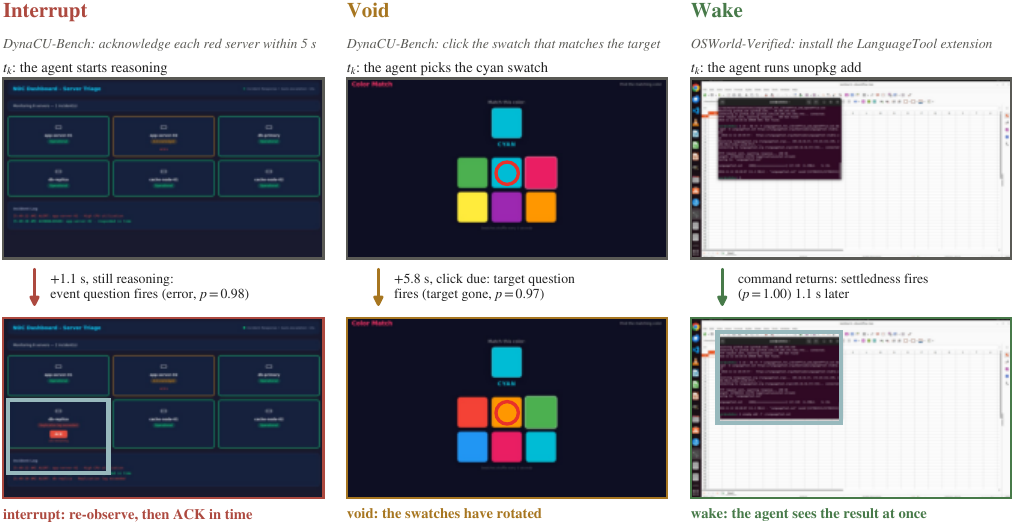}
\caption{\textbf{One intervention of each kind} from the runs of \cref{tab:main} with MiMo-V2.6-Flash. Top: the screenshot the agent reasons on; bottom: the screen when \method intervenes. Interrupt: a second server turns red while the agent is reasoning about the first. Void: the swatches rotate before the click on the cyan one is dispatched (red circle: click point), so the click is discarded. Wake: a terminal command returns, and the agent observes the result 1.1\,s later.}
\label{fig:cases}
\end{figure}

\cref{fig:cases} shows one intervention of each kind.
\textbf{\Int}: on a server dashboard where each alarm must be acknowledged quickly, a second server turns red about 1\,s after the agent began to reason about the previous screen; the error question fires, the generation is cancelled, and the agent acknowledges the new alarm in the restarted step and solves the task.
\textbf{\Void}: in a color-matching game, the agent chooses the cyan swatch, but the swatches rotate during the 5.8\,s it takes to emit the click; the target question fires ($p=0.97$), the click, which would have hit an orange swatch, is discarded, and the agent's next click, chosen on the current screen, solves the task.
\textbf{\Wake}: on OSWorld, the agent installs an extension from a terminal; the watcher finds the terminal settled and wakes the agent 1.1\,s after the command returns, where the time-triggered loop would observe only after its fixed pauses; both methods solve this task in all three runs, and \method spends 222\,s of non-reasoning time per run against 263\,s for Vanilla.
\cref{sec:supp_case} gives further cases.

%% file: sec/5_conclusion.tex
\section{Conclusion}
\label{sec:conclusion}

We presented \method, a plug-and-play framework for proactive perception: the evolving screen, rather than a fixed schedule, decides when an unmodified CUA observes and whether its reasoning and pending action remain valid.
A 0.8B watcher, faster and more successful than a model over $30\times$ its size, answers generic questions about the frame stream, and fixed thresholds turn its answers into interrupts, voids, and wakes.
\method reduces non-reasoning time on OSWorld-Verified by 33\% with task success on par and raises success on DynaCU-Bench from 39.5\% to 44.2\% with the large VLM agent.
These results identify the perception--action gap as a central bottleneck of CUAs and proactive perception as a practical remedy.

\paragraph{Limitations.}
\method changes when the agent perceives, not what it can do: on the 11 OSWorld tasks that fail in every run of both Vanilla and \method, the agent mostly ends the task itself with an incorrect result.
Interrupting and voiding restart the reasoning on the current screen but do not make it faster, so on interfaces that change faster than one generation, such as the fastest toasts and games of DynaCU-Bench, no observation is still valid when the action is dispatched.
\method therefore suits interfaces that change more slowly than the agent reasons, while faster ones need a faster agent (\cref{sec:supp_failure}).

%% file: sec/X_suppl.tex
\newcommand{\imgrole}[1]{$\langle$\textsf{#1}$\rangle$}

\appendix

\section{Question Bank and Hyperparameters}
\label{sec:supp_bank}

\paragraph{Question bank.}
\cref{tab:supp_bank} lists the 16 questions of $\mathcal{Q}$.
Each question is a fixed template: the harness fills in the image roles of \cref{tab:supp_roles} and, where a question refers to the action, a description of the agent's action (\imgrole{action}).
Before dispatch, \imgrole{now} refers to \imgrole{now\_marked}, and after execution, \imgrole{seen} refers to \imgrole{before}.
All questions are yes/no except \texttt{change\_kind}, which chooses among eight options: \texttt{dialog}: a pop-up window, dialog box, permission request or menu; \texttt{toast}: a short-lived notification, toast, tooltip or status message; \texttt{error}: an error or warning message; \texttt{content}: an ordinary content update inside the same page or application; \texttt{navigation}: a different page, window or application; \texttt{keyboard}: an on-screen keyboard or input-method popup; \texttt{ambient}: an ad, carousel, video or animation unrelated to the task; \texttt{trivial}: the mouse cursor, a text caret, a clock or another trivial change.
A change is benign when it is \texttt{ambient} or \texttt{trivial}.
The settledness wake (\cref{eq:wake}) uses only \texttt{settled} and \texttt{loading}, which are served separately from the full checks; a new dialog, error, or toast found by the full check after execution also wakes the agent once it has been drawn.

\begin{table*}[!tbp]
\centering
\caption{\textbf{Question bank.} Text as sent to the watcher, where \imgrole{x} denotes an image role of \cref{tab:supp_roles}. R, D, and E mark the checkpoints that ask the question: during reasoning, before dispatch, and after execution. The answer that fires is ``yes'', except for \texttt{target\_present}, which fires on ``no''.}
\label{tab:supp_bank}
\footnotesize
\begin{tabularx}{\linewidth}{lXccc}
\toprule
\textbf{ID} & \textbf{Question} & \textbf{R} & \textbf{D} & \textbf{E} \\
\midrule
\rowcolor{groupcell}\multicolumn{5}{c}{\textit{A: Change and stability}} \\
\midrule
\texttt{still\_changing} & Compare \imgrole{recent} (taken about half a second earlier) with \imgrole{now}. Is the screen still in the middle of changing: an animation, a transition, scrolling, or content that is still being drawn? &  &  & \ding{51} \\
\texttt{loading} & Does \imgrole{now} show that something is still loading: a spinner, a progress bar, a busy cursor, placeholder blocks, or `Loading' text? &  &  & \ding{51} \\
\texttt{change\_kind} & \imgrole{crop} is a close-up of the part of \imgrole{now} that just changed. What kind of change is it? & \ding{51} &  & \ding{51} \\
\midrule
\rowcolor{groupcell}\multicolumn{5}{c}{\textit{B: Events that make any plan stale}} \\
\midrule
\texttt{dialog} & Compared with \imgrole{seen}, does \imgrole{now} show a new pop-up window, dialog box, permission request or system prompt covering part of the interface? & \ding{51} & \ding{51} & \ding{51} \\
\texttt{error} & Compared with \imgrole{seen}, does \imgrole{now} show a new error, warning or failure message? & \ding{51} &  & \ding{51} \\
\texttt{page\_changed} & Is \imgrole{now} showing a different application, window or page than \imgrole{seen}, rather than the same page with small updates? & \ding{51} &  &  \\
\texttt{toast} & Does \imgrole{now} show a short-lived notification, toast or status message (for example `Saved' or `No network found') that is not in \imgrole{seen}? & \ding{51} &  & \ding{51} \\
\texttt{focus\_lost} & Is the window that is active in \imgrole{seen} no longer active or no longer visible in \imgrole{now}? & \ding{51} &  &  \\
\midrule
\rowcolor{groupcell}\multicolumn{5}{c}{\textit{C: Validity of the pending action}} \\
\midrule
\texttt{target\_present} & \imgrole{target\_seen} shows the element the agent is about to act on, cut from \imgrole{seen} around the action point. \imgrole{target\_now} is the same area of the current screen. Is the same element still there, at the same place, in \imgrole{target\_now}? &  & \ding{51} &  \\
\texttt{target\_occluded} & In \imgrole{now\_marked} the red circle marks where the agent is about to act; the agent chose that point while looking at \imgrole{seen}. Is the point now covered by a pop-up, menu, tooltip, dialog or other overlay that is NOT in \imgrole{seen}? (Clicking a button inside a dialog the agent already saw is not covered.) &  & \ding{51} &  \\
\texttt{target\_changed} & Comparing \imgrole{target\_seen} with \imgrole{target\_now}, did this area change in a way that matters for acting on it (different text, a different element, or a moved layout)? &  & \ding{51} &  \\
\texttt{target\_disabled} & In \imgrole{target\_now}, does the element in the centre look disabled, greyed out, or not yet ready to use? &  & \ding{51} &  \\
\midrule
\rowcolor{groupcell}\multicolumn{5}{c}{\textit{D: Effect of the action and readiness for the next observation}} \\
\midrule
\texttt{effect} & The agent just did this: \imgrole{action}. Comparing \imgrole{before} (what the agent saw before acting) with \imgrole{now}, did the action produce a visible effect on the screen? &  &  & \ding{51} \\
\texttt{settled} & Is \imgrole{now} a settled screen that is ready for the next step, with nothing still loading, animating or half-drawn? &  &  & \ding{51} \\
\texttt{unexpected} & The agent just did this: \imgrole{action}. Does \imgrole{now} show something unexpected as a result, such as the wrong window opening, an error, or the action hitting a different element? &  &  & \ding{51} \\
\texttt{long\_operation} & Does \imgrole{now} show a long operation in progress (saving, installing, downloading, computing) that will take more than a few more seconds? &  &  & \ding{51} \\
\bottomrule
\end{tabularx}
\end{table*}

\begin{table}[!htbp]
\centering
\caption{\textbf{Image roles} of each checkpoint, in the order they are sent to the watcher.}
\label{tab:supp_roles}
\footnotesize
\begin{tabularx}{\linewidth}{lX}
\toprule
\textbf{Role} & \textbf{Description} \\
\midrule
\rowcolor{groupcell}\multicolumn{2}{c}{\textit{During reasoning}} \\
\midrule
\imgrole{seen} & the screenshot the agent is currently reasoning about \\
\imgrole{now} & the current screen \\
\imgrole{crop} & a close-up of the region of the current screen that changed \\
\midrule
\rowcolor{groupcell}\multicolumn{2}{c}{\textit{Before dispatch}} \\
\midrule
\imgrole{seen} & the screenshot the agent based its next action on \\
\imgrole{now\_marked} & the current screen, with a red circle at the point the agent is about to act on \\
\imgrole{target\_seen} & a close-up of the action point in the screenshot the agent saw \\
\imgrole{target\_now} & a close-up of the same area of the current screen \\
\midrule
\rowcolor{groupcell}\multicolumn{2}{c}{\textit{After execution}} \\
\midrule
\imgrole{before} & the screenshot the agent last looked at (its last action was based on it) \\
\imgrole{recent} & the screen about half a second ago \\
\imgrole{now} & the current screen \\
\imgrole{crop} & a close-up of the region changed since the action \\
\bottomrule
\end{tabularx}
\end{table}

\paragraph{Hyperparameters.}
\cref{tab:supp_hparams} lists the thresholds and timing parameters of \cref{sec:interventions}.
They are fixed once and shared across applications, benchmarks, and agents.

\begin{table}[!htbp]
\centering
\caption{\textbf{Hyperparameters} of the intervention rules.}
\label{tab:supp_hparams}
\small
\begin{tabular}{lp{4.6cm}c}
\toprule
\textbf{Symbol} & \textbf{Role} & \textbf{Value} \\
\midrule
$\tau_{I}$ & interrupt on $S_{\text{evt}}$ (dialog, error, page change) & 0.5 \\
$\tau_{V}$ & void on $S_{\text{val}}$ (target gone, occluded, or changed; new dialog) & 0.8 \\
$\tau_{W}$ & wake when the screen is settled & 0.99 \\
$\tau_{L}$ & block waking while loading & 0.5 \\
-- & ignore a change classified as benign & 0.6 \\
\midrule
$t_{\min}$ & minimum quiet time before a wake check & 0.3\,s \\
-- & pixel-quiescence fallback & 0.6\,s \\
-- & no-visible-effect fallback & 1.0\,s \\
-- & maximum wait after an action & 10\,s \\
-- & interrupts per step, at most & 2 \\
-- & voids per step, at most & 2 \\
-- & periodic probe of a static screen & 1.0\,s \\
-- & frame sampling rate & 10\,Hz \\
\bottomrule
\end{tabular}
\end{table}

\paragraph{Wake threshold.}
We evaluated $\tau_{W}$ on the validation split of $\mathcal{T}_{\text{train}}$ (670 examples).
The wake rests on the settledness question, which reaches an AUROC of 0.917 on the validation decision points; the loading question can only postpone a wake, never trigger one.
At 0.3\,s after an action without visible change, $\tau_{W}=0.99$ wakes on 183 of the 196 final screens (93.4\%) and too early at 1 of 201 moments (0.5\%).

\paragraph{Controller.}
The pixel gate compares consecutive frames downscaled to 480\,px grayscale: a pixel changes when it differs by more than 16 gray levels, changed regions smaller than 6 pixels (carets, cursor blinks) and, on OSWorld, the clock in the top bar are ignored, and a change is significant when the remaining changed pixels cover at least 0.04\% of the frame.
An interrupted generation is cancelled and discarded, and the step restarts on the current screen.
A change classified as benign with probability at least 0.6 does not interrupt, and the gate skips its region for the rest of that step.

\section{Prompts and Messages}
\label{sec:supp_prompts}

\tcbset{nwbox/.style={colback=accentlight,colframe=owlbody,boxrule=0.4pt,arc=2pt,left=5pt,right=5pt,top=3pt,bottom=3pt,fontupper=\footnotesize\ttfamily,breakable}}

\paragraph{Watcher request.}
Each request to the watcher consists of a state text, the images of the checkpoint in the order of \cref{tab:supp_roles}, and the questions of the checkpoint (\cref{tab:supp_bank}) as typed fields: yes/no questions are binary, and \texttt{change\_kind} is a choice among the eight options of \cref{sec:supp_bank}.
In the questions, each image role \imgrole{x} is replaced by the name of its image (``image~1'', ``image~2'', \dots).
The state text is
\begin{tcolorbox}[nwbox]
A computer-use agent is operating a desktop. Its task: \textrm{\textit{$\langle$task instruction, at most 240 characters$\rangle$}}\\
Images: image 1 is \textrm{\textit{$\langle$description of the first role$\rangle$}}; image 2 is \textrm{\textit{$\langle$\dots$\rangle$}}.\\
The agent's action: \textrm{\textit{$\langle$action$\rangle$}}.
\end{tcolorbox}
where the last line is present only when there is an action, described as \texttt{click at (x, y)}, \texttt{double click at (x, y)}, \texttt{right click at (x, y)}, \texttt{drag to (x, y)}, \texttt{scroll at (x, y)}, \texttt{type the text '\dots'}, \texttt{press \textrm{\textit{$\langle$keys$\rangle$}}}, or \texttt{run a script}.
The role descriptions are those of \cref{tab:supp_roles}.

\paragraph{Example.}
The request before dispatch in the Void case of \cref{fig:cases}, as sent to the watcher (images omitted):
\begin{tcolorbox}[nwbox]
A computer-use agent is operating a desktop. Its task: A color matching game. A target color is shown at the top. Click the matching color swatch when it appears among rotating options.\\
Images: image 1 is the screenshot the agent based its next action on; image 2 is the current screen, with a red circle at the point the agent is about to act on; image 3 is a close-up of the action point in the screenshot the agent saw; image 4 is a close-up of the same area of the current screen.\\
The agent's action: click at (640, 378).\\[4pt]
dialog (yes/no): Compared with image 1, does image 2 show a new pop-up window, dialog box, permission request or system prompt covering part of the interface?\\
target\_present (yes/no): image 3 shows the element the agent is about to act on, cut from image 1 around the action point. image 4 is the same area of the current screen. Is the same element still there, at the same place, in image 4?\\
target\_occluded (yes/no): In image 2 the red circle marks where the agent is about to act; the agent chose that point while looking at image 1. Is the point now covered by a pop-up, menu, tooltip, dialog or other overlay that is NOT in image 1? (Clicking a button inside a dialog the agent already saw is not covered.)\\
target\_changed (yes/no): Comparing image 3 with image 4, did this area change in a way that matters for acting on it (different text, a different element, or a moved layout)?\\
target\_disabled (yes/no): In image 4, does the element in the centre look disabled, greyed out, or not yet ready to use?
\end{tcolorbox}
The watcher answered ``no'' to \texttt{target\_present} with probability 0.97, above $\tau_{V}=0.8$, so the click was voided.

\paragraph{Teacher.}
The teacher of \cref{sec:training} receives exactly the same requests, with the same state text, images, and questions, so that its answers label the questions that the watcher is later asked.

\paragraph{Messages to the agent.}
\method adds no prompt to the agent and does not change its system prompt.
An interrupt cancels the generation, removes it from the agent's history, and gives the agent the current screenshot as a regular observation, so the agent never sees a partial or discarded answer.
A wake only sets the time of the next observation.
A void on OSWorld is handled like an interrupt, since OSWorld's prompt agent has no channel for failed actions.
On DynaCU-Bench, AOI's agent reports failed actions in its next prompt, and a voided action is reported through the same channel, so that the agent keeps what it read on the old screen:
\begin{tcolorbox}[nwbox]
[ACTION ERROR — previous action failed]\\
\hspace*{1em}Error: not executed: the screen changed before your action could run (\textrm{\textit{$\langle$reason$\rangle$}}). Look at the new screenshot and decide again.\\
\hspace*{1em}Try a different action. Use click(x, y) with pixel coordinates. To type text, use type("text") or click the input first then type("text"). See [PAGE ELEMENTS] above for valid element IDs.
\end{tcolorbox}
where all but the error text is AOI's own feedback, and \textrm{\textit{$\langle$reason$\rangle$}} is the ID of the question that fired.

\section{Watcher Training}
\label{sec:supp_training}

\paragraph{Training tasks.}
$\mathcal{T}_{\text{train}}$ contains 360 tasks over the ten OSWorld application domains: 20 each for Thunderbird and VLC, and 40 for each of the other eight.
An LLM writes the instructions, and each task starts from a freshly launched application on the OSWorld system image, without benchmark files, pages, or profiles.
An instruction is discarded when its string similarity to any OSWorld instruction is at least 0.6, so that no evaluation instruction is reused.

\paragraph{Decision points.}
Running the agent on these tasks yields 344 valid trajectories (16 are skipped because of errors).
They give 14{,}693 training decision points (2{,}540 during reasoning, 4{,}198 before dispatch, and 7{,}955 after execution) and 1{,}512 validation decision points.

\paragraph{Labels.}
We set $\alpha=0.3$ in \cref{eq:labels}, and the teacher is a hosted Jev model.
A gold label is available only where the pixels are unambiguous.

\paragraph{Synthetic interfaces.}
We add 15{,}000 examples from procedurally generated web pages with five layouts and nine event families (dialogs, toasts, errors, loading, navigation, layout shifts, delayed enabling, long operations, and ambient changes), with exact ground truth at every frame.

\paragraph{Settledness examples.}
With a horizon of $h=1$\,s in \cref{eq:hindsight}, the recorded trajectories give 6{,}646 training and 670 validation examples of the moments at which the wake rule decides: 2{,}470 and 263 after a change, and 4{,}176 and 407 after an action that has not changed the screen.

\paragraph{Optimization.}
We fine-tune with LoRA of rank 16 on the attention, MLP, and Gated DeltaNet projection layers, keeping the vision tower frozen and training the readout head in full.
Images are resized to at most 768\,px, $\lambda=1$ in \cref{eq:loss}, and the temperature fitted on the validation split is 0.60.
We use AdamW with a learning rate of $5\times10^{-5}$, linear warmup, and linear decay, and the validation split holds out 10\% of the training tasks.

\section{Experimental Protocol}
\label{sec:supp_protocol}

\paragraph{OSWorld-Verified.}
The sample of 100 tasks contains 9 Chrome, 9 GIMP, 11 Calc, 16 Impress, 8 Writer, 26 multi-application, 8 operating-system, 6 Thunderbird, 5 VLC, and 2 VS~Code tasks.
The screen is $1920\times1080$; the agent receives screenshots downscaled to $1280\times720$, the coordinates in its actions are scaled back to the screen, and it receives the last three screenshots as its history.
The 60\,s environment reset and the 20\,s wait before evaluation are the same for all methods and are excluded from loop time.

\paragraph{DynaCU-Bench.}
Of the 100 dynamic tasks of DynaCU-Bench, we keep the 49 that do not require audio, since audio is disabled for every method: all carousel, transient-notification, and game tasks, 9 of the 10 dashboard and 8 of the 10 video tasks, and one task each of meetings and collaborative editing.
All methods share AOI's harness: AOI's standard agent with its observation text, prompt, action parsing, and failure feedback, with AOI's DOM element list removed before it reaches the agent.
Vanilla and AOI are run as released, and PUSV and \method are added to the same harness, PUSV following its paper.
Every task allows 50 steps instead of the 15 of AOI's evaluation, so that agents are not cut short.
The budget does not drive the results: with MiMo-V2.6-Flash, no successful run of AOI, PUSV, or \method uses more than 15 steps and one of Vanilla does, so a 15-step budget would change success by at most 0.7 points.
Non-reasoning time is reported for PUSV and \method, whose logs separate the agent's loop from environment setup and scoring.
All tasks are scored by AOI's evaluator, as in the original protocol; since the Gemini~2.0~Flash judge of the original evaluation is no longer served, we use Gemini~2.5~Flash and keep the rest of the evaluator unchanged.
PUSV's third layer inspects operating-system alert windows, which a browser page lacks, so here we use its first two layers.

\paragraph{Agents.}
All MiMo-V2.6-Flash calls are routed to a single provider, and all Qwen3.8-27B calls to a single provider's FP8 deployment.
On OSWorld, Qwen3.8-27B decodes greedily, and its system prompt also states the screenshot resolution, since without it the model answers in coordinates of a different resolution.
The scores that the model vendors report on OSWorld-Verified are higher than those of our Vanilla agents, but they come from each vendor's own agent harness and reasoning settings, which are not public.
We therefore run every method with OSWorld's official screenshot-only prompt agent under the same settings, so that the comparison isolates perception timing.

\paragraph{PUSV.}
PUSV checks every click before dispatch, comparing a fresh screenshot with the one the agent reasoned on, with the three layers of its paper~\cite{xu2026toctou}: the SSIM of the $160\times160$ patch at the click falls below 0.92; more than 0.2\% of the pixels outside that patch differ by more than 20 gray levels; or a new window whose title contains an alert keyword appears.
The click is aborted at the first layer that fires, and the agent decides again on the new screen.

\paragraph{Control schemes.}
In \cref{tab:watcher}, the inference delays of D3-27B on the critical path remain in the reported time.

\section{Metric Definitions}
\label{sec:supp_metrics}

All timing is computed from the event log of each episode and the 10\,Hz frame stream, which is the screen's timeline.
A frame-to-frame change is \emph{\textbf{significant}} when the pixel-level gate of \cref{sec:monitoring} fires ($\chi=1$).

\paragraph{Loop and non-reasoning time.}
Loop time runs from the first observation to the end of the episode.
Non-reasoning time is loop time minus all LLM generation, including generations that were interrupted or voided.

\paragraph{Wasted time.}
For each executed action, the time from dispatch to the next screenshot is split at the last significant change before that screenshot.
The part after it, up to the screenshot, is \emph{\textbf{idle}}.
The wasted-time share is the sum of idle time, discarded generations, and the non-idle time of timing-caused no-effect steps, divided by loop time.
A step has no effect when no significant change follows its dispatch, and the no-effect step is timing-caused when the screenshot it acted on was premature or stale.
Thus, the cost $c$ of a premature or stale step in \cref{eq:waste} is measured per step, not set by hand.

\paragraph{Premature and stale.}
Since the validity $v$ in \cref{sec:formulation} cannot be read from the logs, the ledger uses observable proxies.
Both are measured on the screenshot the agent acted on, \ie, the last one before the dispatch, which is a new one when the agent observed again after an interruption or a voided action.
A step is \emph{\textbf{stale}} when the screen changes significantly between that screenshot and the dispatch.
A step is \emph{\textbf{premature}} when that screenshot was taken while the screen was still moving, \ie, significant changes occur both in the 0.5\,s before it and between it and the dispatch.
Both are reported as shares of executed steps.
These proxies count any significant change on the screen; the stricter target-region mismatch below counts only changes at the target and ranks the methods alike (\cref{sec:supp_results}).

\paragraph{Statistical tests.}
\label{sec:supp_stats}
We compare a method with Vanilla per task, averaging its three runs, and test the mean difference over the 100 OSWorld tasks with a paired sign-flip permutation test (20{,}000 permutations).

\paragraph{Mismatch.}
A targeted action is mismatched when at least 5\% of the patch around its target differs between the agent's screenshot and dispatch.

\section{Additional Results}
\label{sec:supp_results}

\cref{tab:supp_osw_domain,tab:supp_dyn_cat} break down the success of MiMo-V2.6-Flash by category (number of tasks in parentheses), although many categories contain few tasks.

\begin{table}[!htbp]
\centering
\caption{\textbf{Success (\%) by domain} on OSWorld-Verified with MiMo-V2.6-Flash (avg@3); parentheses give the number of tasks.}
\label{tab:supp_osw_domain}
\footnotesize
\begin{tabular}{lcccc}
\toprule
\textbf{Category} & \textbf{Vanilla} & \textbf{AOI} & \textbf{PUSV} & \textbf{Ours} \\
\midrule
Chrome (9) & \cellcolor{highlightblue}\textbf{74.1} & \cellcolor{gray!15}70.4 & \cellcolor{highlightblue}\textbf{74.1} & \cellcolor{highlightblue}\textbf{74.1} \\
GIMP (9) & 74.1 & \cellcolor{gray!15}85.2 & 81.5 & \cellcolor{highlightblue}\textbf{92.6} \\
LibreOffice Calc (11) & \cellcolor{gray!15}72.7 & 66.7 & \cellcolor{highlightblue}\textbf{75.8} & \cellcolor{highlightblue}\textbf{75.8} \\
LibreOffice Impress (16) & \cellcolor{gray!15}58.3 & 47.9 & \cellcolor{highlightblue}\textbf{64.6} & 54.2 \\
LibreOffice Writer (8) & \cellcolor{gray!15}75.0 & \cellcolor{gray!15}75.0 & \cellcolor{highlightblue}\textbf{87.5} & \cellcolor{gray!15}75.0 \\
Multi-application (26) & \cellcolor{highlightblue}\textbf{59.0} & \cellcolor{gray!15}53.8 & 52.6 & \cellcolor{highlightblue}\textbf{59.0} \\
Operating system (8) & \cellcolor{highlightblue}\textbf{70.8} & 58.3 & \cellcolor{gray!15}62.5 & \cellcolor{gray!15}62.5 \\
Thunderbird (6) & \cellcolor{gray!15}72.2 & \cellcolor{gray!15}72.2 & \cellcolor{gray!15}72.2 & \cellcolor{highlightblue}\textbf{83.3} \\
VLC (5) & 46.7 & \cellcolor{highlightblue}\textbf{60.0} & \cellcolor{gray!15}53.3 & \cellcolor{gray!15}53.3 \\
VS Code (2) & \cellcolor{highlightblue}\textbf{100.0} & \cellcolor{highlightblue}\textbf{100.0} & \cellcolor{highlightblue}\textbf{100.0} & \cellcolor{highlightblue}\textbf{100.0} \\
\bottomrule
\end{tabular}
\end{table}

\begin{table}[!htbp]
\centering
\caption{\textbf{Success (\%) by category} on DynaCU-Bench with MiMo-V2.6-Flash (avg@3); parentheses give the number of tasks.}
\label{tab:supp_dyn_cat}
\footnotesize
\begin{tabular}{lcccc}
\toprule
\textbf{Category} & \textbf{Vanilla} & \textbf{AOI} & \textbf{PUSV} & \textbf{Ours} \\
\midrule
Meetings (1) & 33.3 & \cellcolor{highlightblue}\textbf{100.0} & \cellcolor{gray!15}66.7 & \cellcolor{gray!15}66.7 \\
Videos (8) & \cellcolor{highlightblue}\textbf{37.5} & 29.2 & \cellcolor{gray!15}33.3 & \cellcolor{highlightblue}\textbf{37.5} \\
Carousels (10) & 23.3 & 23.3 & \cellcolor{gray!15}26.7 & \cellcolor{highlightblue}\textbf{30.0} \\
Dashboards (9) & \cellcolor{gray!15}70.4 & 66.7 & 51.9 & \cellcolor{highlightblue}\textbf{81.5} \\
Transient notif. (10) & \cellcolor{highlightblue}\textbf{36.7} & \cellcolor{highlightblue}\textbf{36.7} & 20.0 & \cellcolor{gray!15}33.3 \\
Collab. editing (1) & \cellcolor{highlightblue}\textbf{100.0} & \cellcolor{gray!15}66.7 & \cellcolor{highlightblue}\textbf{100.0} & \cellcolor{highlightblue}\textbf{100.0} \\
Games (10) & \cellcolor{gray!15}26.7 & 23.3 & \cellcolor{gray!15}26.7 & \cellcolor{highlightblue}\textbf{33.3} \\
\bottomrule
\end{tabular}
\end{table}

\paragraph{Target-region mismatch.}
Under the stricter proxy of \cref{sec:supp_metrics}, which counts only changes at the action's target, 0.8\% of the targeted actions of \method are mismatched on OSWorld against 1.2\% for Vanilla with MiMo-V2.6-Flash, and 0.6\% against 1.1\% with Qwen3.8-27B.

\paragraph{Intervention counts.}
On OSWorld, \method interrupts 0.48 times per task with either agent, mostly on page changes, errors, and dialogs, and voids 0.03 actions per task.
On DynaCU-Bench with MiMo-V2.6-Flash, it interrupts 2.10 times and voids 0.76 actions per task, mostly on page changes and missing targets.
With at most two interrupts and two voids per step and a maximum wait of 10\,s, a page that keeps changing cannot stall a step.

\paragraph{Pixel-only variant on OSWorld.}
To separate the learned watcher from the rest of the harness, we compare \method with a pixel-only variant on OSWorld-Verified under the setting of \cref{tab:main}, with MiMo-V2.6-Flash and one run.
The variant uses the same harness and handles the agent's actions in the same way, but wakes the agent on pixel quiescence alone and never interrupts or voids.
Since \method interrupts only 0.48 times and voids 0.03 actions per task on OSWorld, the difference comes mainly from the wake.
\cref{tab:supp_pixel} compares the time per task: \method spends 37\% less time outside reasoning and 21\% less in total, and is faster on two thirds of the tasks.
On DynaCU-Bench, w/o wake in \cref{tab:interventions} keeps interrupts and voids and replaces only the wake, which adds 9.3\,s per task.

\begin{table}[!htbp]
\centering
\caption{\textbf{\method against a pixel-only variant} on OSWorld-Verified with MiMo-V2.6-Flash (one run, s per task).}
\label{tab:supp_pixel}
\small
\begin{tabular}{lcc}
\toprule
\textbf{Wake rule} & \textbf{Non-reas.}$\downarrow$ & \textbf{Total}$\downarrow$ \\
\midrule
Pixel quiescence & 179.8 & 770.8 \\
\textbf{\watcher (ours)} & \textbf{113.1}\gain{37} & \textbf{611.5}\gain{21} \\
\bottomrule
\end{tabular}
\end{table}

\paragraph{Overheads of passive watching.}
With MiMo-V2.6-Flash on OSWorld, AOI adds a fixed 0.5\,s buffer after every action, and its execution and waiting take 7.5\,s per step against 5.6\,s for Vanilla.
PUSV takes a fresh screenshot and queries the window list before every click, which costs 1.4\,s per checked click, and each of the 3.5 clicks it aborts per 100 steps needs a new decision.

\paragraph{AOI without its DOM list.}
We disable AOI's DOM list so that all methods receive the same screenshot-only input, which isolates perception timing.
Without it, AOI matches Vanilla on carousels and transient notifications and is lower on videos (\cref{tab:supp_dyn_cat}), since its keyframe gates fire rarely on these pages, with 0.19 and 0.33 keyframes per step on videos and transient notifications.

\paragraph{API cost.}
The API cost in \cref{tab:cost} prices the tokens of all agent LLM calls, including interrupted ones, at the list price of each backbone's provider; locally run models, such as our watcher and AOI's CLIP encoder, are not included.

\section{Additional Cases}
\label{sec:supp_case}

Beyond the cases of \cref{fig:cases}, we describe one more interrupt and one more void on DynaCU-Bench.

\paragraph{Interrupt.}
On a DynaCU-Bench page that shows a sequence of error toasts and asks for the third error code, the agent observes the screen 6.5\,s after the start of the task, when only two codes have appeared.
A screenshot-based plan formed on this frame cannot contain the answer.
At 7.9\,s, while the agent is still reasoning, a new error toast appears, the error question fires, and the watcher interrupts the generation, so the next observation, at 8.4\,s, shows the third code, ERR-7712.
A second interrupt at 10.9\,s follows the next toast, and the agent then submits ERR-7712.
\method solves this task in all three runs, and Vanilla in two of them.

\paragraph{Void.}
In a DynaCU-Bench whack-a-mole game, the agent sees a mole in the center hole at 4.9\,s and generates a click on it, at $(640,350)$.
By 9.7\,s, when the action is about to be dispatched, the mole has moved, so the watcher reports that the target is no longer present and the action is discarded.
The next observation, at 9.8\,s, shows the mole in the lower-left hole, and the agent clicks it at $(480,510)$, scoring its first point.
A stale click at the center would have hit an empty hole.
\method solves this task in all three runs, whereas Vanilla fails in all three.

\section{Failure Analysis and Scope}
\label{sec:supp_failure}

\paragraph{Failures caused by the framework.}
A task that Vanilla solves in all three runs and \method in none would point to a failure introduced by the framework.
Among the 149 tasks of the two benchmarks there is one such task, on OSWorld, and conversely two OSWorld tasks are solved in all runs of \method and in none of Vanilla.

\paragraph{Insufficient agent capability.}
On OSWorld, 11 of the 100 tasks fail in all runs of both Vanilla and \method, 6 of them multi-application tasks.
Timing is not what limits them: in 27 of the 33 runs of Vanilla and 26 of the 33 runs of \method the agent ends the task itself with an incorrect result, and 6 runs of each method reach the step limit.
On these tasks \method lowers premature observations from 5.9\% to 3.5\% of steps.
Because \method does not change the agent, it cannot supply a capability that the agent lacks.

\paragraph{Reasoning latency of the agent.}
On DynaCU-Bench, 21 of the 49 tasks fail in all runs of both methods: 6 transient-notification, 5 carousel, 5 game, 4 video, and 1 dashboard task.
Here, 53 of the 63 runs of \method reach the task's time limit, and in the other 10 the agent ends the task with an incorrect result.
The agent spends 9.1\,s of LLM generation per executed step on these tasks, while the toast, frame, or game state that it needs often changes faster than it can respond.
Interrupting and voiding restart the reasoning on the current screen but do not make it faster, so when the screen changes faster than one generation, no observation is still valid when the action is dispatched.
\method therefore suits interfaces that change more slowly than the agent reasons, while faster ones need a faster agent.

\paragraph{Time efficiency of waking.}
Complementing the examples in \cref{sec:supp_case}, the wake intervention shortens the loop whether or not the outcome of a task changes.
On OSWorld with MiMo-V2.6-Flash, execution and waiting together take 3.17\,s per step, against 5.58\,s for Vanilla.

\section{Analysis of Streaming Wake}
\label{sec:supp_wake}

This section estimates how much time the wake rule of \cref{sec:interventions} can save over a fixed delay.
We analyze wake alone, since the stale term of \cref{eq:waste} does not depend on the delay.

\paragraph{Fixed delay.}
Let $T_k=\trdy{k}-\tact{k}$ be the settling duration, with cumulative distribution function $G(x)=\Pr(T_k\le x)$ and mean $\mu$.
A fixed delay $t_{\text{fix}}$ wastes, per step, the expected idle time $\mathbb{E}[(t_{\text{fix}}-T_k)^+]=\int_0^{t_{\text{fix}}}G(x)\,dx$ plus the cost of a premature observation:
\begin{equation}
    W_{\text{fix}}=\int_0^{t_{\text{fix}}}G(x)\,dx+c\,\bigl(1-G(t_{\text{fix}})\bigr).
    \label{eq:wfix}
\end{equation}

\paragraph{Streaming wake.}
A watcher cannot wake at $\trdy{k}$ itself, because settledness is only known in hindsight.
The first frame after $\trdy{k}$ arrives $U\in[0,\epsilon)$ later for a frame interval $\epsilon$, the screen must then stay unchanged across several frames for a quiet time $t_{\text{quiet}}$, and the model answers after a latency $\ell$.
The watcher thus wakes at $\trdy{k}+D$ with $D=U+t_{\text{quiet}}+\ell$ when it confirms settledness at once; a missed confirmation or a loading report delays the wake further, at most to the maximum wait, and we fold this delay into $D$.
The detection delay is then $d=\mathbb{E}[D]\ge\epsilon/2+t_{\text{quiet}}+\ell$, with equality when every confirmation succeeds at once.
With $p_{\text{pre}}$ the probability of waking before $\trdy{k}$, $W_{\text{str}}\le d+c\,p_{\text{pre}}$, and therefore
\begin{equation}
    W_{\text{fix}}-W_{\text{str}}\ge\int_0^{t_{\text{fix}}}G(x)\,dx+c\,\bigl(1-G(t_{\text{fix}})-p_{\text{pre}}\bigr)-d.
    \label{eq:saving}
\end{equation}
A delay that is never premature, $t_{\text{fix}}\ge\max T_k$, saves at least $t_{\text{fix}}-\mu-d-c\,p_{\text{pre}}$ per step: the fixed schedule pays for the spread of settling times, the watcher only for its detection delay.
Shortening $t_{\text{fix}}$ trades idle time for premature observations, and if $T_k$ were constant a fixed delay would be optimal and the watcher would lose $d$.
A shorter quiet time reduces $d$ but raises $p_{\text{pre}}$.

\paragraph{Variability of settling times.}
In the OSWorld runs of Vanilla (MiMo-V2.6-Flash), settling durations have a mean of 2.14\,s, a median of 1.22\,s, and a 99th percentile of 15.8\,s, so no single delay suits all steps.
In the same runs, 9.6\% of the executed steps of Vanilla and 2.6\% of those of \method are premature (\cref{sec:supp_metrics}), so \method lowers idle time and premature observations together.